\documentclass[11pt,a4paper]{article}
\usepackage[hyperref]{acl2020}
\usepackage{times}
\usepackage{latexsym}
\usepackage{graphicx}
\usepackage{multirow}
\usepackage{lscape}

\usepackage{microtype}

\aclfinalcopy 
\def\aclpaperid{1678} 

\title{Automated Topical Component Extraction Using Neural Network Attention Scores from Source-based Essay Scoring}

\author{Haoran Zhang \\
  Department of Computer Science\\
  University of Pittsburgh \\
  Pittsburgh, PA 15260 \\
  \texttt{colinzhang@cs.pitt.edu} \\\And
  Diane Litman \\
  Department of Computer Science \& LRDC\\
  University of Pittsburgh \\
  Pittsburgh, PA 15260 \\
  \texttt{litman@cs.pitt.edu} \\}

\date{}

\begin{document}
\maketitle
\begin{abstract}
While  automated essay scoring (AES) can reliably grade essays at scale, 
automated writing evaluation  (AWE) additionally provides formative feedback  to guide essay revision. However,  
a neural AES typically does not provide useful 
feature representations for supporting AWE.
This paper presents  a method for linking AWE and  neural AES, by extracting Topical Components (TCs) representing evidence from a source text  using the intermediate output of attention layers.
We  evaluate  performance   using a feature-based AES  
requiring TCs.  Results show that performance is comparable whether using automatically  or manually constructed TCs for 1) representing essays as rubric-based features, 2) grading essays.


\end{abstract}

\section{Introduction}
\label{sec:intro}

{\it Automated essay scoring (AES)} systems reliably grade essays at scale, while {\it automated writing evaluation  (AWE}) systems additionally provide formative feedback to guide revision. Although neural networks currently generate state-of-the-art AES results \cite{alikaniotis2016automatic,taghipour2016neural,dong2017attention,farag2018neural,jin2018tdnn,li2018coherence,tay2018skipflow,zhang2018co}, non-neural AES create feature representations more easily useable by AWE \cite{roscoe2014writing,foltz2015analysis,crossley2016adaptive,woods2017formative,madnani2018writing,zhang2019erevise}.
We believe that neural AES can also provide useful information for creating feature representations, e.g., by exploiting information in the intermediate layers.

Our work focuses on a particular source-based essay writing task called the response-to-text assessment (RTA) \cite{correnti2013assessing}.
Recently, an RTA AWE system \cite{zhang2019erevise} was built by extracting rubric-based features related to the use of {\it Topical Components (TCs)} in an essay. However, manual expert effort was first required to create the TCs. For each source, the TCs consist of a comprehensive list of topics related to evidence which include: 1) important words indicating the set of evidence topics in the source, and 2) phrases representing specific examples for each topic that students need to find and use in their essays. 

To eliminate this expert effort, we propose a method for using the interpretable output of the attention layers of a neural AES for source-based essay writing, with the goal of extracting TCs. We evaluate this method by using the  extracted TCs to support feature-based AES for two RTA source texts. Our results show that 
1) the feature-based AES with TCs manually created by humans is matched by our neural method for generating TCs
, and 2) the values of the rubric-based essay features based on automatic  TCs are highly correlated with human Evidence scores.



\begin{table*}[t]
\centering
\scalebox{0.75}{
\begin{tabular}{|p{20cm}|}
\hline {\bf Source Excerpt: }Today, Yala Sub-District {\bf Hospital has medicine}, {\bf free of charge}, {\bf for all of the most common diseases}. {\bf Water is connected to the hospital}, which also has a {\bf generator for electricity}. {\bf Bed nets are used} in every sleeping site in Sauri... \\ \hline
{\bf Essay Prompt: } The author provided one specific example of how the quality of life can be improved by the Millennium Villages Project in Sauri, Kenya. Based on the article, did the author provide a convincing argument that winning the fight against poverty is achievable in our lifetime? Explain why or why not with 3-4 examples from the text to support your answer. \\ \hline
{\bf Essay: }In my opinion I think that they will {\bf achieve it in lifetime}. During the years threw {\bf 2004 and 2008 they made progress}. People didn’t have the money to buy the stuff in 2004. {\bf The hospital was packed with patients} and they didn’t have alot of treatment in 2004. In 2008 it changed the {\bf hospital had medicine}, {\bf free of charge}, and {\bf for all the common dieases}. {\bf Water was connected to the hospital} and has a {\bf generator for electricity}. {\bf Everybody has net} in their site. {\bf The hunger crisis has been addressed} with {\bf fertilizer and seeds}, as well as the {\bf tools needed to maintain the food}. {\bf The school has no fees} and {\bf they serve lunch}. To me that’s sounds like it is going achieve it in the lifetime. \\
\hline
\end{tabular}}
\caption{\label{tab:excerpt} A source excerpt for the $RTA_{MVP}$ prompt and an essay with score of 3.}
\end{table*}

\section{Related Work}

Three recent AWE systems have used non-neural AES to provide rubric-specific feedback.
\citet{woods2017formative} developed an influence estimation process that used a logistic regresion AES to identify sentences needing feedback.
\citet{shibani2019contextualizable} presented a web-based tool that provides formative feedback on rhetorical moves in writing.
\citet{zhang2019erevise} used features created for a random forest AES to select feedback messages, although human effort was first needed to create TCs from a source text.  We automatically extract  TCs using neural AES, thereby eliminating this expert effort.

Others have also proposed methods for pre-processing source information external to an essay.  
Content importance models for AES predict the parts of a source text that students should include when writing a summary~\cite{klebanov2014content}.
Methods for extracting important keywords or keyphrases also exist, both supervised (unlike our approach) \cite{meng2017deep,mahata2018key2vec,florescu2018learning} and unsupervised \cite{florescu2017positionrank}. \citet{rahimi2016automatically} developed a TC extraction LDA model  \cite{blei2003latent}. While the LDA model considers all words equally, our model takes essay scores into account by using attention to represent word importance. Both the unsupervised keyword and LDA models will serve as baselines in our experiments.

In the computer vision area, attention cropped images have been used for further image classification or object detection~\cite{cao2015look,yuxin2018object,ebrahimpour2019ventral}. In the NLP area, \citet{lei2016rationalizing} proposed to use a generator to find candidate rationale and these are passed through the encoder for prediction. Our work is similar in spirit to this type of work.


\section{RTA Corpus and  Prior AES Systems}

\begin{table}[t]
\begin{center}
\scalebox{0.8}{
\begin{tabular}{|r|cc|}
\hline \bf Prompt & \bf $RTA_{MVP}$ & \bf $RTA_{Space}$  \\ \hline
\bf  Score 1 & 852 & 538 \\
& (29\%) & (26\%) \\
\bf Score 2 & 1197 & 789 \\
& (40\%) & (38\%) \\
\bf Score 3 & 616 & 512 \\
& (21\%) & (25\%) \\
\bf Score 4 & 305 & 237 \\
& (10\%) & (11\%) \\ \hline
\bf Total & 2970 & 2076 \\
\hline
\end{tabular}
}
\end{center}
\caption{\label{tab:distribution} The Evidence score distribution of RTA.}
\end{table}

The essays in our corpus were written by students in grades
4 to 8 in response to two RTA source texts  \cite{correnti2013assessing}:  $RTA_{MVP}$ (2970 essays)
and $RTA_{Space}$ (2076 essays).
Table~\ref{tab:excerpt} shows an excerpt from  $RTA_{MVP}$, the associated essay writing prompt, and a student essay. 
The bolding in the source indicates evidence examples  that experts manually labeled as important for students to discuss (i.e., TC phrases). 
Evidence usage in each essay was manually scored on a scale of 1 to 4 (low to high). 
The distribution of Evidence scores is shown in Table~\ref{tab:distribution}.
The essay in Table~\ref{tab:excerpt} received a score of 3, with the bolding indicating  phrases semantically related to the TCs from the source text.

To date, two approaches to AES have been proposed for the RTA: $AES_{rubric}$ and $AES_{neural}$. To support the needs of AWE, 
$AES_{rubric}$~\cite{zhang2017word} used  a traditional supervised learning framework where rubric-motivated  features  were extracted from every essay before model training - Number of Pieces of Evidence (NPE)~\footnote{An integer feature based on the list of {\it topic words} for each topic.}, Concentration (CON), Specificity (SPC)~\footnote{A vector of integer values indicating the number of {\it specific example phrases} (semantically) mentioned in the essay per topic.}, Word Count (WOC).
The two aspects of TCs introduced in Section~\ref{sec:intro} ({\it topic words}, {\it specific example phrases}) were used during feature extraction.








Motivated by improving stand-alone AES performance (i.e., when an interpretable model was not needed for subsequent AWE),  \citet{zhang2018co} developed $AES_{neural}$, a hierarchical neural  model with the co-attention mechanism in the sentence level to capture the relationship between the essay and the source.  Neither feature engineering nor TC creation were needed before  training.   

\section{Attention-Based TC Extraction: $TC_{attn}$}


In this section we propose a method for extracting TCs based on the  $AES_{neural}$ attention level outputs. Since  the self-attention and co-attention mechanisms  were designed to capture  sentence and phrase importance, we hypothesize that the attention scores can help determine if a sentence or phrase has important source-related information.  

To provide intuition, Table~\ref{tab:sents} shows  examples sentences from the student essay in Table~\ref{tab:excerpt}. Bolded are phrases with the highest self-attention score within the sentence. Italics are specific example phrases that refer to the manually constructed TCs for the source. $Attn_{sent}$ is the text to essay attention score that measures which essay sentences have the closest meaning to a source sentence.  $Attn_{phrase}$ is the self-attention score of the bolded phrase that measures phrase importance. 
A sentence with a high attention score tends to include at least one specific example phrase, and vice versa. The phrase with the highest attention score tends to include at least one specific example phrase if the sentence has a high attention score.

\begin{table}[t]
\begin{center}
\scalebox{0.8}{
\small
\begin{tabular}{|p{0.04\linewidth}|p{0.6\linewidth}|p{0.16\linewidth}|p{0.16\linewidth}|}
\hline \bf No. & \bf Sentences & {\bf $attn_{sent}$} & {\bf $attn_{phrase}$}  \\ \hline
1 & People didn't {\bf have the money to buy} the stuff in 2004. & 0.00420 & 0.23372\\ \hline
2 & The \textit{hunger crisis has been} {\bf \textit{addressed with fertilizer and seeds}}, as well as the \textit{tools needed to maintain the food}. & 0.08709 & 0.62848 \\ \hline
3 & {\bf \textit{The school has no fees}} and \textit{they serve lunch}.  & 0.10686 & 0.63369 \\ \hline

\end{tabular}
}
\end{center}
\caption{\label{tab:sents} Example attention scores of essay sentences.}
\end{table}

Based on these observations, we first extract the output of two layers from the neural network: 1) the $attn_{sent}$ of each sentence, and 2) the output of the convolutional layer as the representation of the phrase with the highest $attn_{phrase}$ in each sentence (denoted by $cnn_{phrase}$). We also extract the plain text of the phrase with the highest $attn_{phrase}$ in each sentence (denoted by $text_{phrase}$).
Then, our  $TC_{attn}$ method uses the extracted information in  3 main steps: 1) filtering out $text_{phrase}$ from sentences with low $attn_{sent}$, 2) clustering all remaining $text_{phrase}$ based on $cnn_{phrase}$, and 3) generating TCs from clusters.

The first filtering step  keeps all $text_{phrase}$ where the original sentences have $attn_{sent}$  higher than a threshold. The intuition is that  lower $attn_{sent}$ indicates less source-related information.

The second step clusters these $text_{phrase}$ based on their corresponding representations $cnn_{phrase}$. We use k-medoids  to cluster $text_{phrase}$ into   $M$ clusters, where  $M$ is the number of topics in the source text. Then, for  $text_{phrase}$ in each topic cluster, we use  k-medoids  to cluster them into  $N$ clusters, where $N$ is the number of the specific example phrases we want to extract from each topic. The outputs of this step are $M*N$ clusters.

The third step uses  the topic and example clustering to extract TCs. As noted earlier, TCs include two parts: topic words, and specific example phrases. Since our method is data-driven  and  students  introduce their vocabulary into the corpus, essay text is  noisy. To make the TC output cleaner, we  filter out words that are not in the source text. To obtain topic words, we combine all $text_{phrase}$ from each topic cluster to calculate the word frequency per topic. To make 
topics unique, we 
assign each word to the topic cluster in which it has the highest normalized word frequency. We then include the top $K_{topic}$ words based on their frequency in each topic cluster. To obtain  example phrases, we combine all $text_{phrase}$ from each example cluster to calculate the word frequency per example, then include the top $K_{example}$ words based on their frequency in each example cluster. 

\section{Experimental Setup and Results}
\label{sec:exp}

\begin{table}[t]
\begin{center}
\scalebox{0.8}{
\begin{tabular}{|c|c|c|}
\hline \bf Layer & \bf Parameter Name & \bf Value  \\ \hline
Embedding & Embedding dimension & 50 \\ \hline
Word-CNN & Kernel size & 5 \\
& Number of filters & 100 \\ \hline
Sent-LSTM & Hidden units & 100 \\ \hline
Modeling & Hidden units & 100 \\ \hline
Dropout & Dropout rate & 0.5 \\ \hline
Others & Epochs & 100 \\
& Batch size & 100 \\
& Initial learning rate & 0.001 \\
& Momentum & 0.9 \\
\hline
\end{tabular}}
\end{center}
\caption{\label{tab:hyper} Hyper-parameters for neural training.}
\end{table}

\begin{figure}[t]
\centering
\includegraphics[height=.12\textheight]{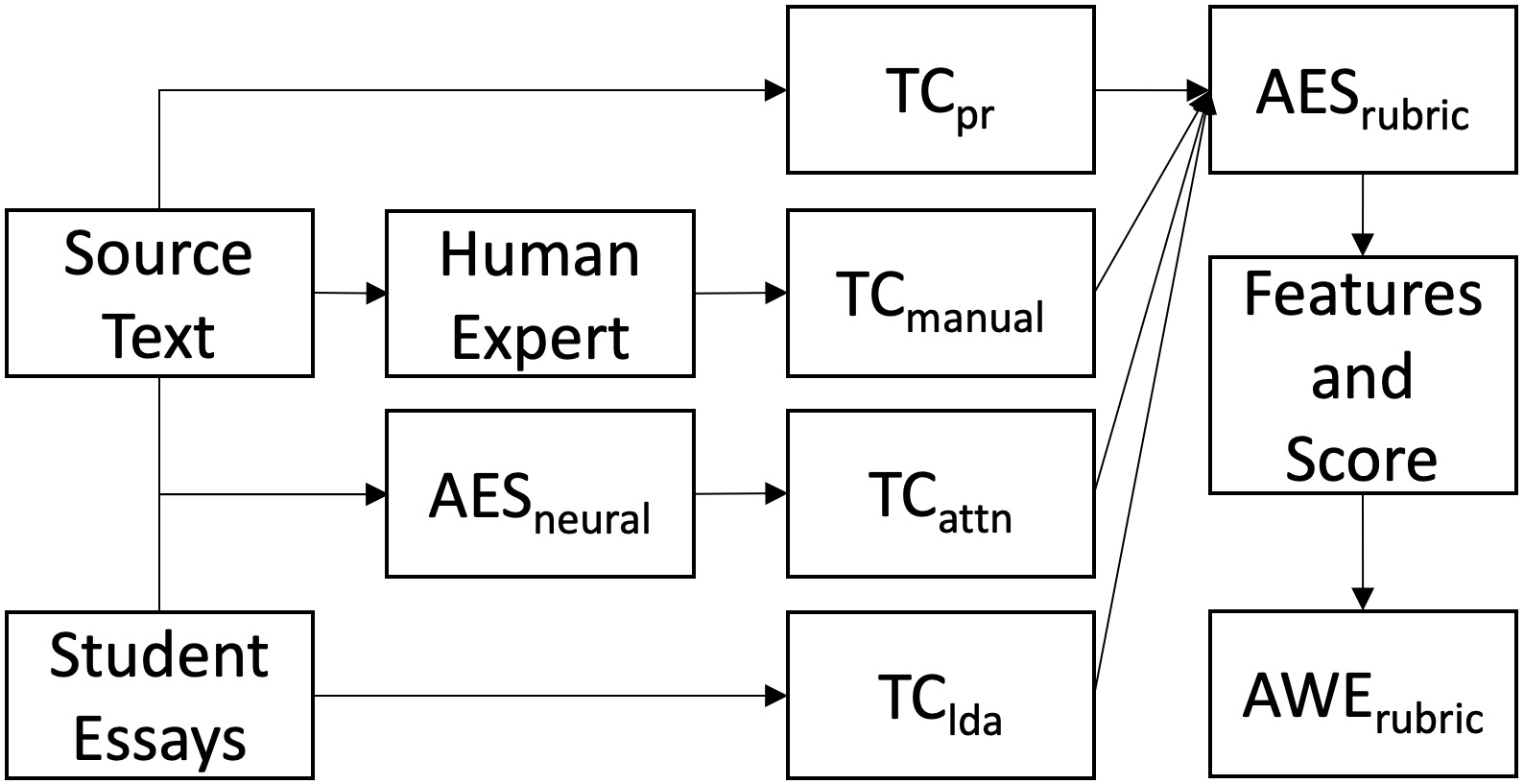}
\caption{An overview of four TC extraction systems.}
\label{fig:overview}
\end{figure}

Figure~\ref{fig:overview} shows an overview of  four TC extraction methods to be evaluated. $TC_{manual}$ (upper bound) uses a human expert to extract TCs  from a source text.
$TC_{attn}$ is our proposed method and automatically extracts TCs using {\it both} a source text and student essays.  
$TC_{lda}$ \cite{rahimi2016automatically} (baseline) builds on LDA to extract TCs from student essays only, while $TC_{pr}$ (baseline) builds on PositionRank~\cite{florescu2017positionrank} to instead extract TCs from only the source text.

Since PositionRank is not designed for TC extraction, we needed to further process its output to create $TC_{pr}$. To extract topic words, we extract all keywords from the output. Next, we map each word to a higher dimension with word embedding. Lastly, we cluster all keywords using k-medoids  into $PR_{topic}$ topics. To extract example phrases, we put them into only one topic and remove all redundant example phrases if they are subsets of other example phrases.

We configure experiments to test two hypotheses: H1) the $AES_{rubric}$  model for scoring Evidence \cite{zhang2017word} will perform comparably when extracting features using either $TC_{attn}$ or $TC_{manual}$, and will perform worse when using $TC_{lda}$ or $TC_{pr}$; H2) the correlation between the human Evidence score and the feature values (NPE and sum of SPC features)\footnote{These features are extracted based on TCs.} 
will be comparable when extracted using $TC_{attn}$ and $TC_{manual}$, and will be stronger than when using $TC_{lda}$ and $TC_{pr}$.
The experiment for H1 tests the impact of using our proposed TC extraction method on the downstream $AES_{rubric}$ task, while the H2 experiment examines the impact on the essay representation itself.

Following \citet{zhang2017word}, we stratify essay corpora: 40\%  for training word embeddings and extracting TCs, 20\% for selecting the best  embedding and parameters, and  40\% for testing. We use the hyper-parameters from \citet{zhang2018co} for neural  training as shown in Table~\ref{tab:hyper}. Table~\ref{tab:parameters} shows all other parameters selected using the development set.

\begin{table}[t]
\begin{center}
\scalebox{0.6}{
\begin{tabular}{l|l|l|ccc}
Prompt & Component & Parameter & $TC_{lda}$ & $TC_{pr}$ & $TC_{attn}$ \\
\hline
\multirow{4}{*}{$RTA_{MVP}$} & \multirow{2}{*}{Topic Words} & Number of Topics & 9 & 19 & 16 \\
 & & Number of Words & 30 & 20 & 25 \\
\cline{2-6}
 & \multirow{2}{*}{Example Phrases} & Number of Topics & 20 & 1 & 18 \\
 & & Number of Phrases & 15 & 20 & 15 \\

\hline
\multirow{4}{*}{$RTA_{Space}$} & \multirow{2}{*}{Topic Words} & Number of Topics & 15 & 20 & 10 \\
 & & Number of Words & 10 & 10 & 20 \\
\cline{2-6}
 & \multirow{2}{*}{Example Phrases} & Number of Topics & 10 & 1 & 9 \\
 & & Number of Phrases & 20 & 50 & 20 \\

\end{tabular}}
\end{center}
\caption{\label{tab:parameters} Parameters for different models.}
\end{table}

\begin{table}[t]
\begin{center}
\scalebox{0.68}{
\begin{tabular}{l|llll}
Prompt & $TC_{manual}$ (1) & $TC_{lda}$ (2) & $TC_{pr}$ (3) & $TC_{attn}$ (4) \\ \hline
$RTA_{MVP}$ & 0.643 (2,3) & 0.614 (3) & 0.525 & \bf 0.648 (1,2,3) \\
$RTA_{Space}$ & 0.609 (3) & 0.615 (3) & 0.559 &\bf 0.622 (1,3) \\

\end{tabular}}
\end{center}
\caption{\label{tab:result-qwk} The performance (QWK) of $AES_{rubric}$ using different TC extraction methods for feature creation. The numbers in the parentheses show the model numbers over which the current model performs significantly better ($p<0.05$). The best results between automated methods in each row are in bold.}
\end{table}


{\bf Results for H1.} H1 is supported by the results in Table~\ref{tab:result-qwk}, which compares the Quadratic Weighted Kappa (QWK) between human and $AES_{rubric}$ Evidence scores (values 1-4) when $AES_{rubric}$ uses $TC_{manual}$ versus each of the automatic methods. $TC_{attn}$ always yields better performance, and even significantly better than $TC_{manual}$. 

{\bf Results for H2.} The results  in Table~\ref{tab:result-feature} support H2. $TC_{attn}$ outperforms the two automated baselines, and for NPE even yields stronger correlations than the manual TC method.


\begin{table}[t]
\begin{center}
\scalebox{0.7}{
\begin{tabular}{ll|cccc}
\hline Prompt & Feature & $TC_{manual}$ & $TC_{lda}$ & $TC_{pr}$ & $TC_{attn}$ \\ \hline
\multirow{2}{*}{$RTA_{MVP}$} & NPE & 0.542 & 0.482 & 0.587 & \bf 0.639 \\
& SPC (sum) & 0.689 & 0.585 & 0.365 & \bf 0.679 \\
\hline
\multirow{2}{*}{$RTA_{Space}$} & NPE & 0.484 & 0.513 & 0.494 & \bf 0.625 \\
& SPC (sum) & 0.601 & 0.574 & 0.533 & \bf 0.598 \\
\hline
\end{tabular}}
\end{center}
\caption{\label{tab:result-feature}Pearson's r comparing feature values computed using each TC extraction method with human (gold-standard) Evidence essay scores. All correlation values are significant ($p\leq0.05$). The best results between automated methods in each row are in bold.}
\end{table}

\begin{table*}[t]
\begin{center}
\scalebox{0.6}{
\begin{tabular}{|cccc|}
\hline $TC_{manual}$ & $TC_{lda}$ & $TC_{pr}$ & $TC_{attn}$ \\ \hline
progress just four years & running water electricity & brighter future hannah & electricity running water irrigation set \\
medicine most common diseases & water connected hospital generator electricity & millennium villages project & poor showed treatment school supplies \\
water connected hospital & patients afford & unpaved dirt road & farmers could crops afford bed \\
hospital generator electricity & rooms packed patients probably & bar sauri primary school & electricity hospital \\
bed nets used every sleeping site & share beds & future hannah & better fertilizer medicine enough also \\
hunger crisis addressed fertilizer seeds & recieve treatment & sauri primary school & rooms packed patients \\
tools needed maintain food supply & doctor clinical officer running hospital & villages project & food fertilizer crops get supply \\
no school fees & doctors clinical & millennium development goals & five net costs 5 \\
school attendance rate way up & water fertilizer knowledge & village leaders & nets net bed free \\
kids go school now & receive treatment & dirt road & running water supplies schools almost \\
... & ... & ... & ... \\
\hline
\end{tabular}}
\end{center}
\caption{\label{tab:result-example}Specific example phrases for the $RTA_{MVP}$ progress topic.}
\end{table*}

{\bf Qualitative Analysis.}
The manually-created topic words for $RTA_{MVP}$ represent 4 topics, which are ``hospital'', ``malaria'', ``farming'' and ``school''\footnote{All Topic Words generated by different models can be found in the Appendix \ref{app:tw}.}. Although Table~\ref{tab:parameters} shows that the automated list has more topics for topic words and might have broken one topic into separate topics, a good automated list should have more topics related to the 4 topics above. We manually assign a topic for each of the topic words from the different automated methods. $TC_{lda}$ has 4 related topics out of 9 (44.44\%), $TC_{pr}$ has 6 related topics out of 19 (31.58\%), and $TC_{attn}$ has 10 related topics out of 16 (62.50\%). Obviously, $TC_{attn}$ preserves more related topics than our baselines.

Moving to the second aspect of TCs (specific example phrases), Table~\ref{tab:result-example} shows the first 10 specific example phrases for a manually-created category that introduces the changes made by the MVP project\footnote{All Specific Example Phrases generated by different models can be found in the Appendix \ref{app:sep}.}. This category is a mixture of different topics because it talks about the ``hospital'', ``malaria'', ``school'', and ``farming'' at the same time. $TC_{attn}$ has overlap with $TC_{manual}$ on different topics. However, $TC_{lda}$ mainly talks about ``hospital'', because the nature of the LDA model doesn't allow mixing specific example phrases about different topics in one category. Unfortunately, $TC_{pr}$ does not include any overlapped specific phrase in the first 10 items; they all refer to some general example phrases from the beginning of the source article. Although there are some related specific example phrases in the full list, they are mainly about school. This is because the PositionRank algorithm tends to assign higher scores to words that appear early in the text.

\section{Conclusion and Future Work}
This paper proposes $TC_{attn}$, a method for  using the attention scores in a neural AES model 
to automatically extract the Topical Components of a source text. Evaluations show the potential of $TC_{attn}$ for eliminating expert effort without degrading $AES_{rubric}$ performance or the feature representations themselves. $TC_{attn}$   outperforms  baselines and  generates comparable or even better results than a manual approach.

Although $TC_{attn}$ outperforms all baselines and requires no human effort on TC extraction, annotation of essay evidence scores is still needed. This leads to an interesting future investigation direction, which is training the $AES_{neural}$ using the gold standard that can be extracted automatically. 

One of our next steps is to investigate the impact of TC extraction methods on a corresponding AWE system~\cite{zhang2019erevise}, which uses the feature values produced by $AES_{rubric}$ to generate formative feedback to guide essay revision. 

Currently, the $TC_{lda}$ are trained on student essays, while the $TC_{pr}$ only works on the source article. However, $TC_{attn}$ uses both student essays and the source article for TC generation. It might be hard to say that the superior performance of $TC_{attn}$ is due to the neural architecture and attention scores rather than the richer training resources. Therefore, a comparison between $TC_{attn}$ and a model that uses both student essays and the source article is needed.

\section*{Acknowledgments}
We would like to show our appreciation to every member of the RTA group for sharing their pearls of wisdom with us. We are also immensely grateful to all members of the PETAL group and reviewers for their comments on an earlier version of the paper.

The research reported here was supported, in whole or in part, by the Institute of Education Sciences, U.S. Department of Education, through Grant R305A160245 to the University of Pittsburgh. The opinions expressed are those of the authors and do not represent the views of the Institute or the U.S. Department of Education.

\bibliography{anthology,acl2020}

\begin{thebibliography}{28}
\expandafter\ifx\csname natexlab\endcsname\relax\def\natexlab#1{#1}\fi

\bibitem[{Alikaniotis et~al.(2016)Alikaniotis, Yannakoudakis, and
  Rei}]{alikaniotis2016automatic}
Dimitrios Alikaniotis, Helen Yannakoudakis, and Marek Rei. 2016.
\newblock Automatic text scoring using neural networks.
\newblock In \emph{Proceedings of the 54th Annual Meeting of the Association
  for Computational Linguistics (Volume 1: Long Papers)}, volume~1, pages
  715--725.

\bibitem[{Blei et~al.(2003)Blei, Ng, and Jordan}]{blei2003latent}
David~M Blei, Andrew~Y Ng, and Michael~I Jordan. 2003.
\newblock Latent dirichlet allocation.
\newblock \emph{Journal of machine Learning research}, 3(Jan):993--1022.

\bibitem[{Cao et~al.(2015)Cao, Liu, Yang, Yu, Wang, Wang, Huang, Wang, Huang,
  Xu et~al.}]{cao2015look}
Chunshui Cao, Xianming Liu, Yi~Yang, Yinan Yu, Jiang Wang, Zilei Wang, Yongzhen
  Huang, Liang Wang, Chang Huang, Wei Xu, et~al. 2015.
\newblock Look and think twice: Capturing top-down visual attention with
  feedback convolutional neural networks.
\newblock In \emph{Proceedings of the IEEE International Conference on Computer
  Vision}, pages 2956--2964.

\bibitem[{Correnti et~al.(2013)Correnti, Matsumura, Hamilton, and
  Wang}]{correnti2013assessing}
Richard Correnti, Lindsay~Clare Matsumura, Laura Hamilton, and Elaine Wang.
  2013.
\newblock Assessing students' skills at writing analytically in response to
  texts.
\newblock \emph{The Elementary School Journal}, 114(2):142--177.

\bibitem[{Crossley and McNamara(2016)}]{crossley2016adaptive}
Scott~A Crossley and Danielle~S McNamara. 2016.
\newblock \emph{Adaptive educational technologies for literacy instruction}.
\newblock Routledge.

\bibitem[{Dong et~al.(2017)Dong, Zhang, and Yang}]{dong2017attention}
Fei Dong, Yue Zhang, and Jie Yang. 2017.
\newblock Attention-based recurrent convolutional neural network for automatic
  essay scoring.
\newblock In \emph{Proceedings of the 21st Conference on Computational Natural
  Language Learning (CoNLL 2017)}, pages 153--162.

\bibitem[{Ebrahimpour et~al.(2019)Ebrahimpour, Li, Yu, Reesee, Moghtaderi,
  Yang, and Noelle}]{ebrahimpour2019ventral}
Mohammad~K Ebrahimpour, Jiayun Li, Yen-Yun Yu, Jackson Reesee, Azadeh
  Moghtaderi, Ming-Hsuan Yang, and David~C Noelle. 2019.
\newblock Ventral-dorsal neural networks: Object detection via selective
  attention.
\newblock In \emph{2019 IEEE Winter Conference on Applications of Computer
  Vision (WACV)}, pages 986--994. IEEE.

\bibitem[{Farag et~al.(2018)Farag, Yannakoudakis, and
  Briscoe}]{farag2018neural}
Youmna Farag, Helen Yannakoudakis, and Ted Briscoe. 2018.
\newblock Neural automated essay scoring and coherence modeling for
  adversarially crafted input.
\newblock In \emph{Proceedings of the 2018 Conference of the North American
  Chapter of the Association for Computational Linguistics: Human Language
  Technologies, Volume 1 (Long Papers)}, pages 263--271.

\bibitem[{Florescu and Caragea(2017)}]{florescu2017positionrank}
Corina Florescu and Cornelia Caragea. 2017.
\newblock Positionrank: An unsupervised approach to keyphrase extraction from
  scholarly documents.
\newblock In \emph{Proceedings of the 55th Annual Meeting of the Association
  for Computational Linguistics (Volume 1: Long Papers)}, pages 1105--1115.

\bibitem[{Florescu and Jin(2018)}]{florescu2018learning}
Corina Florescu and Wei Jin. 2018.
\newblock Learning feature representations for keyphrase extraction.
\newblock In \emph{Thirty-Second AAAI Conference on Artificial Intelligence}.

\bibitem[{Foltz and Rosenstein(2015)}]{foltz2015analysis}
Peter~W Foltz and Mark Rosenstein. 2015.
\newblock Analysis of a large-scale formative writing assessment system with
  automated feedback.
\newblock In \emph{Proceedings of the Second (2015) ACM Conference on Learning@
  Scale}, pages 339--342. ACM.

\bibitem[{Jin et~al.(2018)Jin, He, Hui, and Sun}]{jin2018tdnn}
Cancan Jin, Ben He, Kai Hui, and Le~Sun. 2018.
\newblock Tdnn: a two-stage deep neural network for prompt-independent
  automated essay scoring.
\newblock In \emph{Proceedings of the 56th Annual Meeting of the Association
  for Computational Linguistics (Volume 1: Long Papers)}, pages 1088--1097.

\bibitem[{Klebanov et~al.(2014)Klebanov, Madnani, Burstein, and
  Somasundaran}]{klebanov2014content}
Beata~Beigman Klebanov, Nitin Madnani, Jill Burstein, and Swapna Somasundaran.
  2014.
\newblock Content importance models for scoring writing from sources.
\newblock In \emph{Proceedings of the 52nd Annual Meeting of the Association
  for Computational Linguistics (Volume 2: Short Papers)}, pages 247--252.

\bibitem[{Lei et~al.(2016)Lei, Barzilay, and Jaakkola}]{lei2016rationalizing}
Tao Lei, Regina Barzilay, and Tommi Jaakkola. 2016.
\newblock Rationalizing neural predictions.
\newblock In \emph{Proceedings of the 2016 Conference on Empirical Methods in
  Natural Language Processing}, pages 107--117.

\bibitem[{Li et~al.(2018)Li, Chen, Nie, Liu, Feng, and Cai}]{li2018coherence}
Xia Li, Minping Chen, Jianyun Nie, Zhenxing Liu, Ziheng Feng, and Yingdan Cai.
  2018.
\newblock Coherence-based automated essay scoring using self-attention.
\newblock In \emph{Chinese Computational Linguistics and Natural Language
  Processing Based on Naturally Annotated Big Data}, pages 386--397. Springer.

\bibitem[{Madnani et~al.(2018)Madnani, Burstein, Elliot, Klebanov, Napolitano,
  Andreyev, and Schwartz}]{madnani2018writing}
Nitin Madnani, Jill Burstein, Norbert Elliot, Beata~Beigman Klebanov, Diane
  Napolitano, Slava Andreyev, and Maxwell Schwartz. 2018.
\newblock Writing mentor: Self-regulated writing feedback for struggling
  writers.
\newblock In \emph{Proceedings of the 27th International Conference on
  Computational Linguistics: System Demonstrations}, pages 113--117.

\bibitem[{Mahata et~al.(2018)Mahata, Kuriakose, Shah, and
  Zimmermann}]{mahata2018key2vec}
Debanjan Mahata, John Kuriakose, Rajiv~Ratn Shah, and Roger Zimmermann. 2018.
\newblock Key2vec: Automatic ranked keyphrase extraction from scientific
  articles using phrase embeddings.
\newblock In \emph{Proceedings of the 2018 Conference of the North American
  Chapter of the Association for Computational Linguistics: Human Language
  Technologies, Volume 2 (Short Papers)}, pages 634--639.

\bibitem[{Meng et~al.(2017)Meng, Zhao, Han, He, Brusilovsky, and
  Chi}]{meng2017deep}
Rui Meng, Sanqiang Zhao, Shuguang Han, Daqing He, Peter Brusilovsky, and
  Yu~Chi. 2017.
\newblock Deep keyphrase generation.
\newblock In \emph{Proceedings of the 55th Annual Meeting of the Association
  for Computational Linguistics (Volume 1: Long Papers)}, pages 582--592.

\bibitem[{Rahimi and Litman(2016)}]{rahimi2016automatically}
Zahra Rahimi and Diane Litman. 2016.
\newblock Automatically extracting topical components for a response-to-text
  writing assessment.
\newblock In \emph{Proceedings of the 11th Workshop on Innovative Use of NLP
  for Building Educational Applications}, pages 277--282.

\bibitem[{Roscoe et~al.(2014)Roscoe, Allen, Weston, Crossley, and
  McNamara}]{roscoe2014writing}
Rod~D Roscoe, Laura~K Allen, Jennifer~L Weston, Scott~A Crossley, and
  Danielle~S McNamara. 2014.
\newblock The writing pal intelligent tutoring system: Usability testing and
  development.
\newblock \emph{Computers and Composition}, 34:39--59.

\bibitem[{Shibani et~al.(2019)Shibani, Knight, and
  Shum}]{shibani2019contextualizable}
Antonette Shibani, Simon Knight, and Simon~Buckingham Shum. 2019.
\newblock Contextualizable learning analytics design: A generic model and
  writing analytics evaluations.
\newblock In \emph{Proceedings of the 9th International Conference on Learning
  Analytics \& Knowledge}, pages 210--219. ACM.

\bibitem[{Taghipour and Ng(2016)}]{taghipour2016neural}
Kaveh Taghipour and Hwee~Tou Ng. 2016.
\newblock A neural approach to automated essay scoring.
\newblock In \emph{Proceedings of the 2016 Conference on Empirical Methods in
  Natural Language Processing}, pages 1882--1891.

\bibitem[{Tay et~al.(2018)Tay, Phan, Tuan, and Hui}]{tay2018skipflow}
Yi~Tay, Minh~C Phan, Luu~Anh Tuan, and Siu~Cheung Hui. 2018.
\newblock Skipflow: incorporating neural coherence features for end-to-end
  automatic text scoring.
\newblock In \emph{Thirty-Second AAAI Conference on Artificial Intelligence}.

\bibitem[{Woods et~al.(2017)Woods, Adamson, Miel, and
  Mayfield}]{woods2017formative}
Bronwyn Woods, David Adamson, Shayne Miel, and Elijah Mayfield. 2017.
\newblock Formative essay feedback using predictive scoring models.
\newblock In \emph{Proceedings of the 23rd ACM SIGKDD International Conference
  on Knowledge Discovery and Data Mining}, pages 2071--2080. ACM.

\bibitem[{Yuxin et~al.(2018)Yuxin, Xiangteng, and Junjie}]{yuxin2018object}
Peng Yuxin, He~Xiangteng, and Zhao Junjie. 2018.
\newblock Object-part attention model for fine-grained image classification.
\newblock \emph{IEEE transactions on image processing: a publication of the
  IEEE Signal Processing Society}, 27(3):1487--1500.

\bibitem[{Zhang and Litman(2017)}]{zhang2017word}
Haoran Zhang and Diane Litman. 2017.
\newblock Word embedding for response-to-text assessment of evidence.
\newblock In \emph{Proceedings of ACL 2017, Student Research Workshop}, pages
  75--81.

\bibitem[{Zhang and Litman(2018)}]{zhang2018co}
Haoran Zhang and Diane Litman. 2018.
\newblock Co-attention based neural network for source-dependent essay scoring.
\newblock In \emph{Proceedings of the Thirteenth Workshop on Innovative Use of
  NLP for Building Educational Applications}, pages 399--409.

\bibitem[{Zhang et~al.(2019)Zhang, Magooda, Litman, Correnti, Wang, Matsmura,
  Howe, and Quintana}]{zhang2019erevise}
Haoran Zhang, Ahmed Magooda, Diane Litman, Richard Correnti, Elaine Wang,
  LC~Matsmura, Emily Howe, and Rafael Quintana. 2019.
\newblock erevise: Using natural language processing to provide formative
  feedback on text evidence usage in student writing.
\newblock In \emph{Proceedings of the AAAI Conference on Artificial
  Intelligence}, volume~33, pages 9619--9625.

\end{thebibliography}
\bibliographystyle{acl_natbib}

\clearpage

\appendix

\section{Appendices}

\subsection{Topic Words Results}
\label{app:tw}
Table~\ref{tab:topic-manual} shows all topic words for the $RTA_{MVP}$ from $TC_{manual}$. Table~\ref{tab:topic-lda} shows all topic words for the $RTA_{MVP}$ from $TC_{lda}$. Table~\ref{tab:topic-pr} shows all topic words for the $RTA_{MVP}$ from $TC_{pr}$. Table~\ref{tab:topic-attn} shows all topic words for the $RTA_{MVP}$ from $TC_{attn}$.

\subsection{Specific Example Phrases Results}
\label{app:sep}
Table~\ref{tab:example-manual} shows all specific example phrases for the $RTA_{MVP}$ from $TC_{manual}$. Table~\ref{tab:example-lda} shows all specific example phrases for the $RTA_{MVP}$ from $TC_{lda}$. Table~\ref{tab:example-pr} shows all specific example phrases for the $RTA_{MVP}$ from $TC_{pr}$. Table~\ref{tab:example-attn} shows all specific example phrases for the $RTA_{MVP}$ from $TC_{attn}$.

\begin{table*}[t]
\begin{center}
\begin{tabular}{|cccc|}
\hline Topic 1 & Topic 2 & Topic 3 & Topic 4 \\ \hline
care & bed & farmer & school\\
health & net & fertilizer & supplies\\
hospital & malaria & irrigation & fee\\
treatment & infect & dying & student\\
doctor & bednet & crop & midday\\
electricity & mosquito & seed & meal\\
disease & bug & water & lunch\\
water & sleeping & harvest & supply\\
sick & die & hungry & book\\
medicine & cheap & feed & paper\\
generator & infect & food & pencil\\
no & biting &  & energy\\
die &  &  & free\\
kid &  &  & children\\
bed &  &  & kid\\
patient &  &  & go\\
clinical &  &  & attend\\
officer &  &  & \\
running &  &  & \\
\hline
\end{tabular}
\end{center}
\caption{\label{tab:topic-manual}Topic words of $TC_{manual}$.}
\end{table*}

\begin{landscape}
\begin{table}[t]
\begin{center}
\begin{tabular}{|ccccccccc|}
\hline Topic 1 & Topic 2 & Topic 3 & Topic 4 & Topic 5 & Topic 6 & Topic 7 & Topic 8 & Topic 9 \\ \hline
help & kenya & poverty & food & money & school & people & hospital & years \\
poor & like & think & fertilizer & need & kids & sauri & medicine & africa \\
world & better & author & crops & nets & supplies & malaria & hospitals & project \\
good & know & lifetime & water & thing & children & sick & water & villages \\
things & life & article & farmers & afford & schools & 2008 & free & sauri \\
time & help & possible & needed & donate & lunch & disease & electricity & village \\
work & think & convinced & grow & right & education & 2004 & diseases & helped \\
hard & sauri & fight & dying & dollar & afford & nets & medicines & change \\
going & live & proverty & problem & treatment & energy & mosquitoes & doctors & lives \\
alot & clothes & said & family & survive & learn & getting & 2008 & goals \\
reason & states & achievable & families & needs & students & says & gave & improved \\
happen & place & time & stop & stuff & went & years & doctor & 2015 \\
helping & health & convince & lack & person & adults & progress & examples & help \\
goal & important & believe & hunger & cause & fees & died & 2004 & changed \\
believe & feel & hannah & tools & patients & parents & text & shape & year \\
problems & happy & shows & seeds & provide & 2004 & away & cure & changes \\
countries & tell & reasons & plants & cost & lunches & mosquitos & running & started \\
difference & care & convincing & fertilizers & beds & books & prevent & treat & great \\
places & shoes & fighting & farming & means & home & treated & support & millennium \\
change & story & wrote & able & dont & wanted & dieing & common & progress \\
little & america & story & solved & dollars & chores & said & beds & came \\
improve & ways & agree & supply & medical & meal & come & patients & girl \\
country & wants & saying & irrigation & jobs & wood & night & said & 2025 \\
achieve & makes & opinion & wont & everyday & materials & bite & generator & place \\
hope & clothing & winning & afford & gone & learning & death & clean & program \\
helps & community & sachs & hungry & doctors & able & sleep & electricty & tells \\
everybody & economy & progress & plant & lots & suplies & impoverished & giving & small \\
start & history & conclusion & look & sickness & meals & living & drink & millenium \\
easy & paragraph & says & farms & live & paper & amazing & cures & read \\
making & thats & future & feed & fact & attendance & easily & evidence & happened \\
\hline
\end{tabular}
\end{center}
\caption{\label{tab:topic-lda}Topic words of $TC_{lda}$.}
\end{table}
\end{landscape}

\begin{landscape}
\begin{table}[t]
\begin{center}
\scalebox{0.65}{
\begin{tabular}{|ccccccccccccccccccc|}
\hline Topic 1 & Topic 2 & Topic 3 & Topic 4 & Topic 5 & Topic 6 & Topic 7 & Topic 8 & Topic 9 & Topic 10 & Topic 11 & Topic 12 & Topic 13 & Topic 14 & Topic 15 & Topic 16 & Topic 17 & Topic 18 & Topic 19 \\ \hline
irrigation & road & diseases & adults & fight & development & joyful & people & midday & village & millennium & backs & plenty & doctor & thing & paper & end & work & sleeping \\
fertilizer & brighter & medicine &  & lifetime & villages & dirt & kids & school &  &  & women & access & hospital &  & supplies &  & world & bed \\
farmers & future & malaria &  &  & project & jump &  & fees &  &  & ground & care & shape &  & chores &  &  & net \\
crops & hannah & disease &  &  & goals & bar &  & students &  &  & bananas & medicines & patients &  & books &  &  & nets \\
plant & car & mosquitoes &  &  & plan & music &  & meal &  &  & cloth & schools & treatment &  & pencils &  &  & site \\
seeds & sauri & charge &  &  & economy & singing &  & energy &  &  & mothers & today & officer &  &  &  &  &  \\
outcome & market &  &  &  & quality & everyone &  & lunch &  &  & feet & supply & water &  &  &  &  &  \\
lack & year &  &  &  & supporters & dancing &  &  &  &  & clothing & areas & electricity &  &  &  &  &  \\
tools & time &  &  &  &  & help &  &  &  &  & day & kind & generator &  &  &  &  &  \\
 & place &  &  &  &  & health &  &  &  &  & rooms &  &  &  &  &  &  &  \\
 & years &  &  &  &  & advice &  &  &  &  & family &  &  &  &  &  &  &  \\
 & poverty &  &  &  &  & items &  &  &  &  &  &  &  &  &  &  &  &  \\
 & life &  &  &  &  & targets &  &  &  &  &  &  &  &  &  &  &  &  \\
 & communities &  &  &  &  & death &  &  &  &  &  &  &  &  &  &  &  &  \\
 & leaders &  &  &  &  & night &  &  &  &  &  &  &  &  &  &  &  &  \\
 & glimpse &  &  &  &  & costs &  &  &  &  &  &  &  &  &  &  &  &  \\
 & africa &  &  &  &  & die &  &  &  &  &  &  &  &  &  &  &  &  \\
 & chemicals &  &  &  &  & knowledge &  &  &  &  &  &  &  &  &  &  &  &  \\
 & solutions &  &  &  &  & food &  &  &  &  &  &  &  &  &  &  &  &  \\
 & millions &  &  &  &  & parents &  &  &  &  &  &  &  &  &  &  &  & \\
\hline
\end{tabular}
}
\end{center}
\caption{\label{tab:topic-pr}Topic words of $TC_{pr}$.}
\end{table}
\end{landscape}

\begin{landscape}
\begin{table}[t]
\begin{center}
\scalebox{0.75}{
\begin{tabular}{|cccccccccccccccc|}
\hline Topic 1 & Topic 2 & Topic 3 & Topic 4 & Topic 5 & Topic 6 & Topic 7 & Topic 8 & Topic 9 & Topic 10 & Topic 11 & Topic 12 & Topic 13 & Topic 14 & Topic 15 & Topic 16 \\ \hline
poverty & way & years & lunch & goals & electricity & supplies & afford & many & free & school & hospital & bed & project & supply & fertilizer \\
fight & would & four & serves & problems & water & food & lifetime & people & medicine & schools & 2004 & nets & world & maintain & seeds \\
winning & rate & villages & parents & day & generator & net & could & kenya & crops & fees & disease & used & millennium & diseases & addressed \\
 & attendance & 80 & attend & cloth & also & rooms & achievable & sauri & charge & students & yala & every & village & hunger & irrigation \\
 & help & progress & passed & three & running & packed & together & pencils & farmers &  &  & sleeping & across & lives & necessary \\
 & kids & last &  & made & energy & patients & malaria & africa & medicines &  &  & site & work & adults & tools \\
 & enough & occurred &  & books & connected & needed & take & yet &  &  &  & midday & end & life & lack \\
 & better & year &  & 2015 &  & 5 & future & sachs &  &  &  & meal & worry & dying & plenty \\
 & go & changes &  & knowledge &  & keep & worked & though &  &  &  & dramatic & supporters & death & plant \\
 & get & outcome &  & learn &  & poor & care & feed &  &  &  & change & time & away & common \\
 & place & today &  & one &  & five & family & two &  &  &  & clinical & 2025 & treated & become \\
 & solutions & first &  &  &  & like & hard & health &  &  &  & officer & history &  &  \\
 & really & along &  &  &  & come & good & set &  &  &  & tattered & selling &  &  \\
 & targets &  &  &  &  & little & doctor & crisis &  &  &  & clothing &  &  &  \\
 & see &  &  &  &  & treatment & either & areas &  &  &  & chemicals &  &  &  \\
 & die &  &  &  &  & minimal & whole & items &  &  &  & malarial &  &  &  \\
 & hungry &  &  &  &  & almost & save &  &  &  &  & preventable &  &  &  \\
 & dancing &  &  &  &  & harvest & millions &  &  &  &  & treatable &  &  &  \\
 & walked &  &  &  &  & showed & easy &  &  &  &  & costs &  &  &  \\
 & bare &  &  &  &  & cheap & met &  &  &  &  &  &  &  &  \\
 & feet &  &  &  &  &  & ever &  &  &  &  &  &  &  &  \\
 & hannah &  &  &  &  &  & around &  &  &  &  &  &  &  &  \\
 & impoverished &  &  &  &  &  & mosquitoes &  &  &  &  &  &  &  &  \\
 & encouraging &  &  &  &  &  & easily &  &  &  &  &  &  &  &  \\
 & probably &  &  &  &  &  &  &  &  &  &  &  &  &  & \\
\hline
\end{tabular}
}
\end{center}
\caption{\label{tab:topic-attn}Topic words of $TC_{attn}$.}
\end{table}
\end{landscape}

\begin{landscape}
\begin{table}[t]
\begin{center}
\scalebox{0.65}{
\centering
\begin{tabular}{|cccc|}
\hline Category 1 & Category 2 & Category 3 & Category 4 \\ \hline
unpaved roads & united nations intervention & yala sub district hospital & malaria common disease preventable treatable \\
tattered clothing & safer healthier better life & three kids bed two adults rooms packed patients & mosquitoes carry malaria infect people biting \\
bare feet & out poverty stabilize economy quality life communities & not medicine treatment could afford & kids die malaria adults sick 20 000 day \\
less than 1 dollar day & africa kenya sauri & no doctor only clinical officer running hospital & bed nets mosquitoes away people save millions lives \\
 & goals met 2015 2025 & no running water electricity & bed nets cost 5 dollar \\
 & 80 villages across sub-sahara africa & sad people dying near death preventable & cheap medicines treat malaria \\
\hline Category 5 & Category 6 & Category 7 & Category 8 \\ \hline
crops dying & kids not attend go school & progress just four years & progress encouraging supporters \\
not afford fertilizer irrigation & not afford school fees & yala sub district hospital has medicine & solutions problems keep people impoverished \\
outcome poor crops & kids help chores fetching water wood & medicine free charge & change poverty stricken areas good \\
lack fertilizer water & schools minimal supplies books paper pencils & medicine most common diseases & poverty history not easy task hard \\
enough food crops harvest feed whole family hungry sick & concentrate not energy & water connected hospital & winning against poverty possible achievable lifetime \\
 & no midday meal lunch & hospital generator electricity &  \\
 &  & bed nets used every sleeping site &  \\
 &  & hunger crisis addressed fertilizer seeds &  \\
 &  & tools needed maintain food supply &  \\
 &  & kids go school now &  \\
 &  & no school fees &  \\
 &  & now serves lunch students &  \\
 &  & school attendance rate way up & \\
\hline
\end{tabular}
}
\end{center}
\caption{\label{tab:example-manual}Specific example phrases of $TC_{manual}$.}

\end{table}
\end{landscape}

\begin{landscape}
\begin{table}[t]
\begin{center}
\scalebox{0.5}{
\centering
\begin{tabular}{|ccccc|}
\hline Category 1 & Category 2 & Category 3 & Category 4 & Category 5 \\ \hline
work hard & life time & author convince winning fight poverty achievable lifetime & children adults & easy task \\
better place & united nations & author convinced winning fight poverty achievable lifetime & mosquitoes carry malaria & lived dollar \\
better health & united states & author wants & disease called malaria & thing history \\
brighter future & life communities & author convince winning fight proverty & come night & stuff need \\
things like & like books paper pencils & winning fight proverty achievable lifetime & malarial mosquitoes & earn money \\
things need & learn life kenya & winning fight poverty achievable life time & easily adults sick &  \\
fighting poverty & important kids & article brighter future & solutions problems people impoverished &  \\
work change & thinks important & wining fight poverty achievable & mosquitoes away &  \\
hard work & wants know & article states & infect people biting &  \\
agree author &  & winning fight poverty acheivable & away sleeping &  \\
working hard &  & author provided &  &  \\
better life 2008 &  & author thinks &  &  \\
better life2008 &  & based article author convince &  &  \\
reading article &  & convinced poverty &  &  \\
things changed &  & poverty acheivable lifetime &  &  \\
\hline Category 6 & Category 7 & Category 8 & Category 9 & Category 10 \\ \hline
attendance rate & amazing progress years & good shape & kids adults & donate money \\
midday meal & text says & good education & 2015 2025 & tattered clothes \\
serves lunch students & text said & went school & hungry sick & tattered clothing \\
midday meals & year girl & areas good & cheap medicines & bare feet \\
served lunch & year 2004 & trying help & goals supposed & donating money \\
students wanted learn & paragraph says & worked hard &  & save millions lives \\
books pencils & progress shows winning fight poverty achievable & second reason &  &  \\
kids attend school & treated chemicals & second example &  &  \\
schools minimal & paragraph states & girl went &  &  \\
schools hospitals & progress encouraging supporters millennium villages & hannah sachs convinced winning &  &  \\
school school fees &  & went kenya &  &  \\
practical items &  &  &  &  \\
kids sauri attend school parents afford school fees &  &  &  &  \\
attendence rate &  &  &  &  \\
parents money &  &  &  &  \\
\hline Category 11 & Category 12 & Category 13 & Category 14 & Category 15 \\ \hline
clean water & grow crops & millennium village project & stop poverty & running water electricity \\
water wood & feed family & millenium village project & long time & water connected hospital generator electricity \\
fresh water & needed help & millennium villages project helped & world work change & patients afford \\
needs help & farmers worry & change dramatically & beat poverty & rooms packed patients probably \\
medicines free charge & crops dying afford necessary fertilizer irrigation & dramatic changes occured villages subsaharan africa & ending poverty & share beds \\
chores fetching & fertilizer knowledge & place live & want learn & recieve treatment \\
fetching water & hunger crisis addressed fertilizer seeds tools needed maintain food supply & happened years & places like & doctor clinical officer running hospital \\
 & feed families & dramatic changes occurred villages & shows winning fight poverty achievable lifetime & doctors clinical \\
 & hunger crisis adressed & millennium development goals & want kind poverty & water fertilizer knowledge \\
 & family plant seeds outcome poor & change povertystricken areas good & poverty assure access & receive treatment \\
 & farmers worried & coming years &  & running bare \\
 &  & encouraging supporters millennium villages project &  & afford treatment \\
 &  & occurred villages subsaharan &  &  \\
\hline Category 16 & Category 17 & Category 18 & Category 19 & Category 20 \\ \hline
yala subdistrict hospital medicine free charge common diseases & nets sleeping site sauri & plan people poverty & achieve goal & years later \\
free lunch & afford nets & stabilize economy quality life communities & reach goal & took years \\
yala district &  & assure access health care help & going school & started 2004 \\
preventable treatable &  & people people & story says &  \\
common africa &  & near death & achieve goals &  \\
diseases like &  & poor crops lack &  &  \\
common disease africa &  & homeless people &  &  \\
hospital good shape &  &  &  &  \\
district hospital &  &  &  & \\
\hline
\end{tabular}
}
\end{center}
\caption{\label{tab:example-lda}Specific example phrases of $TC_{lda}$.}

\end{table}
\end{landscape}

\begin{table*}[t]
\begin{center}
\centering
\begin{tabular}{|c|}
\hline Category 1 \\ \hline
brighter future hannah \\
millennium villages project \\
unpaved dirt road \\
bar sauri primary school \\
future hannah \\
sauri primary school \\
villages project \\
millennium development goals \\
village leaders \\
dirt road \\
car jump \\
little kids \\
preventable diseases people \\
many kids \\
diseases people \\
kids die \\
school supplies \\
primary school \\
school fees \\
infect people \\
\hline
\end{tabular}
\end{center}
\caption{\label{tab:example-pr}Specific example phrases of $TC_{pr}$.}

\end{table*}

\begin{landscape}
\begin{table}[t]
\begin{center}
\scalebox{0.5}{
\centering
\begin{tabular}{|cccccc|}
\hline Category 1 & Category 2 & Category 3 & Category 4 & Category 5 & Category 6 \\ \hline
winning fight & could feed bed net afford & four years progress lifetime year & fees students school supplies schools & sauri knowledge & supplies medicines \\
poverty winning world villages & people school work hard books & villages occurred 80 across along & school fees supplies afford fertilizer & afford school fees & better medicine water energy \\
winning fight poverty & also every diseases kids health & net 5 & tools crops school fees seeds & bed nets help keep & hospital electricity connected \\
winning poverty & preventable family people care & years many villages sauri project & farmers rooms patients crops people & food attendance rooms end many & bed nets 5 also \\
fight poverty & afford school fees bed nets & outcome poor crops & school lunch meal midday supplies & problems also people energy many & water electricity hospital fertilizer \\
poverty fight winning & also would energy learn help & progress years kenya africa today & lunch students serves midday & food supply maintain electricity supplies & electricity water energy \\
fight poverty winning & people fees school farmers could & rate people & medicine 2004 5 years keep & school fees & bed showed \\
 & lunch could work electricity medicine & villages kenya 80 farmers many & school lunch schools also fees & 2004 also year rate school & bed nets used \\
 & could afford fertilizer & four years lifetime poverty year & years school showed hospital water & farmers needed food supply villages & generator energy \\
 & school supplies little afford enough & years four last five day & school parents attend &  & bed nets free \\
 & food also & farmers two many poverty & school medicine fertilizer hospital bed &  & water electricity also fertilizer supplies \\
 & also tools & years changes fertilizer addressed & school schools fees free two &  & electricity water running also generator \\
 & supply maintain food also tattered & years villages kenya project attendance & school fees schools lunch free &  & generator electricity \\
 &  & energy poverty hunger electricity & lunch school crops food farmers &  & fertilizer bed net water \\
 &  &  & water fertilizer energy school medicines &  & fertilizer addressed school supplies crisis \\
\hline Category 7 & Category 8 & Category 9 & Category 10 & Category 11 & Category 12 \\ \hline
electricity running water irrigation set & bed showed diseases & help students supplies people schools & years four free schools medicine & medicine electricity tools fertilizer medicines & schools also school students attendance \\
poor showed treatment school supplies & lunch meal energy & people years four three though & school schools free supplies fees & water electricity connected schools running & free charge school maintain supply \\
farmers could crops afford bed & dramatic change bed nets & villages years 80 poverty many & crops fertilizer farmers tools plant & students lunch serves school 2004 & crops farmers 2004 first food \\
electricity hospital & poverty better lives made many & worked together end & water electricity supplies school energy & medicine crops free hospital also & lack fertilizer school bed nets \\
better fertilizer medicine enough also & achievable lifetime sauri & pencils students supplies yet & medicine school supplies years hunger & school supplies farmers attendance crops & bed nets years hospital \\
rooms packed patients & malaria good bed net used & villages many kenya sauri 80 & fertilizer crops lack farmers water & water supplies schools free hospital & hospital disease four years 2004 \\
food fertilizer crops get supply & bed net & years food supply hunger crisis & fertilizer irrigation crops medicine water & schools crops supplies free charge & every sleeping site \\
five net costs 5 & common diseases & sauri net & medicines school medicine fertilizer free & school schools lunch also free & school bed also occurred 80 \\
nets net bed free & work together poverty & net 5 & free charge medicine school medicines & school fees schools & years four schools last students \\
running water supplies schools almost & hospital go school could afford & school supplies items & seeds plant crops fertilizer & school fees lunch & school supplies schools also 2004 \\
bed supplies knowledge medicines afford & project progress made food good & sachs many & free schools lunch school charge & lunch schools school seeds food & crops farmers schools project also \\
supplies food supply farmers water & also hospital doctor clinical showed &  & bed nets water fertilizer medicines & school fees schools free lunch & hospital years medicine school water \\
supplies midday school food hunger & years made malaria take changes &  & free charge medicine school fertilizer & schools supplies electricity farmers fertilizer & free charge schools years meal \\
many food & could better future people lunch &  & crops farmers fertilizer electricity knowledge & students lunch & medicine hospital made \\
 &  &  & school fees schools free medicines & schools school farmers crops bed & free charge school years hunger \\
\hline Category 13 & Category 14 & Category 15 & Category 16 & Category 17 & Category 18 \\ \hline
bed nets & villages africa millennium 80 across & supply books & seeds fertilizer addressed food medicine & enough would work hard better & water connected hospital \\
water running medicine medicines supplies & 80 villages across & electricity water & seeds supply fertilizer crops plenty & people world sauri kids poverty & nets bed used crops afford \\
bed nets medicine crops electricity & poverty fight people kenya end & poverty many lives hunger every & fertilizer seeds crops & many people poverty could take & midday meal \\
sauri free bed nets & world 2015 & diseases lack water day every & tools fertilizer & kenya would better walked bare & midday meal lunch \\
crops fertilizer plant food irrigation & poor village sauri & adults one bed two last & crops farmers also water could & poverty problems crisis though many & bed nets used \\
bed nets every water medicine & well project villages poor end & people food work many energy & crops seeds water needed & people kenya targets 80 villages & bed every sleeping site net \\
fertilizer crops water keep tools & achievable kenya & villages village school people many & addressed fertilizer seeds & almost kids die people & bed nets every used school \\
kenya bed nets & many villages people problems kenya & school food schools hospital people & seeds fertilizer food also water & rate way progress better africa & hospital water running clinical officer \\
bed nets also adults & project villages kenya village people & years changes four free occurred & seeds fertilizer water & attendance rate way & water hospital bed nets \\
sauri bed nets & goals four years met needed & water every work school fees & fertilizer food & see world & bed nets could keep \\
every bed nets & poverty village fight africa sauri & years hospital villages charge connected & fertilizer irrigation necessary farmers tools & go hungry get people could & bed nets used every sleeping \\
diseases medicine medicines common preventable & attendance rate way selling come & food maintain supply electricity supplies & fertilizer seeds irrigation farmers lack & get food work would probably & hospital charge bed nets preventable \\
nets bed water sauri years & work world help last together & 2015 2025 dying hunger death & fertilizer lack crops become sauri & world winning fight way place &  \\
crops fertilizer enough farmers & poverty many 2015 millennium progress & diseases malaria &  & people easily sauri history way &  \\
 &  & site sauri &  & help people poverty place many & \\
\hline
\end{tabular}
}
\end{center}
\caption{\label{tab:example-attn}Specific example phrases of $TC_{attn}$.}

\end{table}
\end{landscape}

\end{document}